\pdfoutput=1

\documentclass[11pt]{article}

\usepackage[final]{acl}

\usepackage{times}
\usepackage{latexsym}
\usepackage{xcolor}
\usepackage{subcaption}
\usepackage{pdfpages}
\usepackage{caption}
\usepackage{makecell}
\usepackage{booktabs}
\usepackage[T1]{fontenc}

\usepackage[utf8]{inputenc}
\usepackage{longtable}
\newcommand{\scheme}{SlideCoder}

\newcommand{\bench}{Slide2Code}

\newcommand{\model}{SlideMaster}

\usepackage{microtype}

\usepackage{inconsolata}
\usepackage{amsmath}
\usepackage{booktabs}
\usepackage{multirow}
\usepackage{graphicx}
\usepackage{geometry}
\usepackage{caption}
\usepackage{xcolor}
\usepackage{colortbl}
\definecolor{color1}{RGB}{250,255,218}
\definecolor{color11}{RGB}{227,242,217}
\definecolor{color2}{RGB}{255,243,202}
\definecolor{low}{RGB}{252,228,211}
\usepackage{multirow}
\usepackage{graphicx}
\usepackage{algorithm}
\usepackage{algorithmic}
\usepackage{tabularx}
\title{%
  \raisebox{-1.2ex}{\includegraphics[width=1cm]{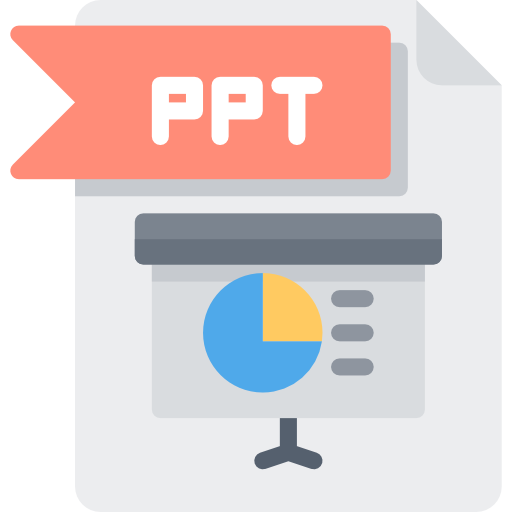}}\hspace{0.4em}
  {\huge\bfseries SlideCoder: Layout‑aware RAG‑enhanced Hierarchical Slide Generation from Design}
}

\title{%
  \raisebox{-1.2ex}{\includegraphics[width=1cm]{figure/ppt.png}}
  {SlideCoder: Layout‑aware RAG‑enhanced Hierarchical Slide Generation from Design}
}

\author{
\normalfont
Wenxin Tang$^{1}$\thanks{These authors contributed equally.},
Jingyu Xiao$^{2}$\footnotemark[1],
Wenxuan Jiang$^{3}$,
Xi Xiao$^{1}$,
Yuhang Wang$^{4}$,\\
Xuxin Tang$^{5}$,
Qing Li$^{6}$,
Yuehe Ma$^{7}$,
Junliang Liu$^{8}$,
Shisong Tang$^{5}$,
Michael R. Lyu$^{2}$\\
$^1$Tsinghua University, $^2$The Chinese University of Hong Kong, $^3$Northeastern University\\ 
$^4$Southwest University, $^5$Kuaishou Technology, $^6$Peng Cheng Laboratory\\
$^7$BNU-HKBU United International College, $^8$Dalian Maritime University\\
twx24@mails.tsinghua.edu.cn, jyxiao@link.cuhk.edu.hk, xiaox@sz.tsinghua.edu.cn\\
}

\begin{document}
\maketitle
\begin{abstract}
Manual slide creation is labor-intensive and requires expert prior knowledge. Existing natural language-based LLM generation methods struggle to capture the visual and structural nuances of slide designs. To address this, we formalize the Reference Image to Slide Generation task and propose Slide2Code, the first benchmark with difficulty-tiered samples based on a novel Slide Complexity Metric. We introduce SlideCoder, a layout-aware, retrieval-augmented framework for generating editable slides from reference images. SlideCoder integrates a Color Gradient-based Segmentation algorithm and a Hierarchical Retrieval-Augmented Generation method to decompose complex tasks and enhance code generation. We also release SlideMaster, a 7B open-source model fine-tuned with improved reverse-engineered data. Experiments show that SlideCoder outperforms state-of-the-art baselines by up to 40.5 points, demonstrating strong performance across layout fidelity, execution accuracy, and visual consistency. Our code is available at \url{https://github.com/vinsontang1/SlideCoder}.

\end{abstract}

\section{Introduction}

Slide creation is essential in academic and professional communication for visually conveying complex ideas. However, manual design is labor-intensive and time-consuming~\cite{oldppt5}. While templates offer some relief, they enforce fixed layouts and styles, limiting flexibility. 


Recent progress in Large Language Models (LLMs)~\cite{llm1,llm2} has sparked interest in automatic slide creation. AutoPresent~\cite{ge2025autopresent}, an early study on the Natural Language (NL) to slide generation task, fine-tunes a LLAMA-based model~\cite{llama} on the diversified SLIDESBENCH dataset. It translates NL instructions into Python code, which invokes SLIDESLIB, a high-level API built on python-pptx~\cite{pythonpptx}, to construct each slide. This pipeline reduces manual effort and streamlines design workflows.

Despite Autopresent's capability to generate slides from natural language input, several significant challenges remain unaddressed. 

\begin{figure}[t]
\small
  \includegraphics[width=\columnwidth]{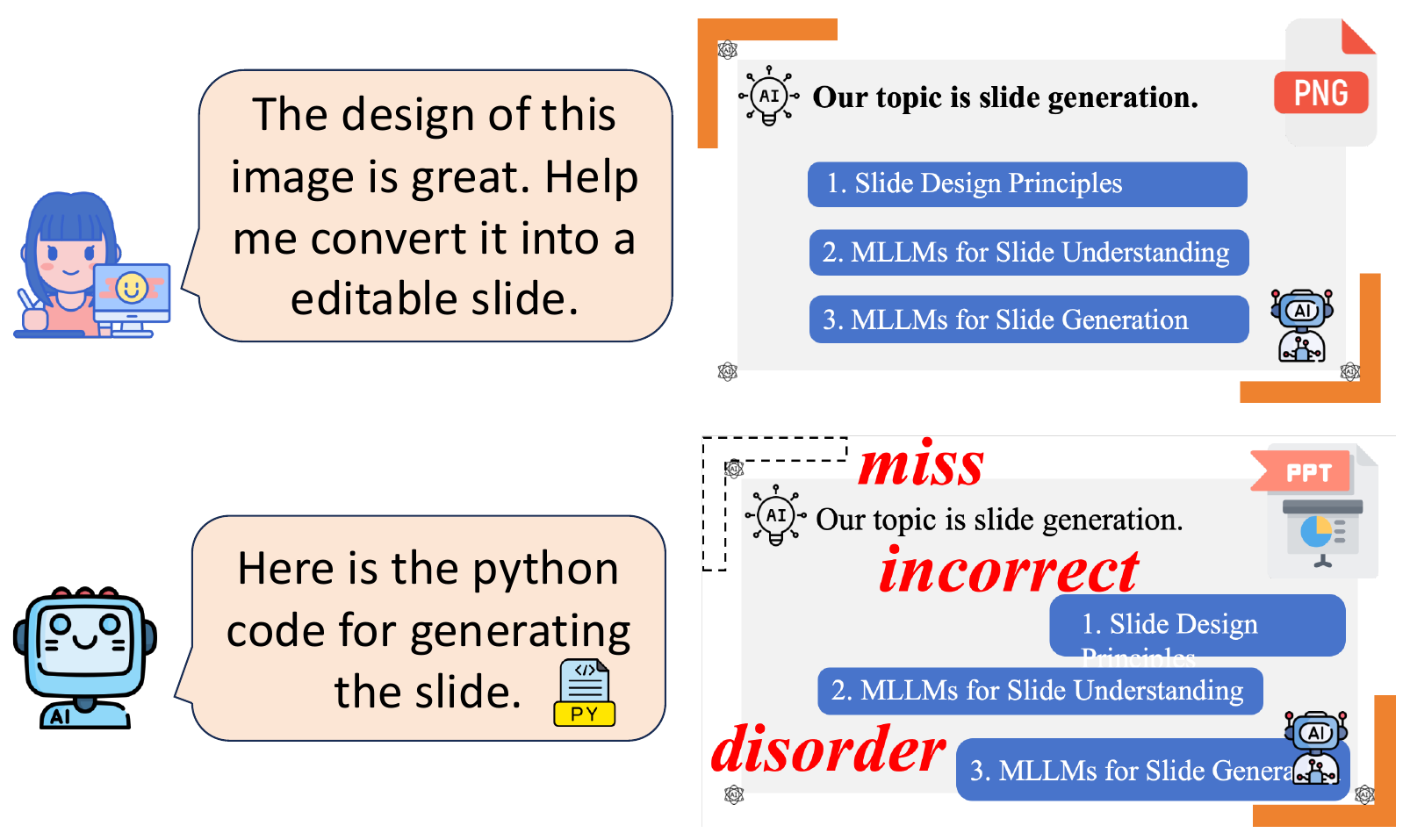}
  \caption{Illustration of slide generation scenarios from design and mistakes made by MLLMs.}
  \label{fig:errortypes}
\end{figure}

\textbf{First, natural language inherently lacks an accurate description of slide visual design (e.g., color, layout, and style) and users sometimes directly input the design image for slide generation.} For example, as shown in Figure~\ref{fig:errortypes}, a user sees a nice design from non-editable slides (png and pdf format) or other source like webpage design, and hopes to convert it into an editable slide (pptx format). Or the user lacks the skills to make slides, they can generate the slide by input their design image. In these scenarios, the Multimodal Large Language Models (MLLMs) are needed to understand the design and generate slides.



\textbf{Second, MLLMs face limitations when handling complex slides, particularly those incorporating diverse element types and high element density.} As illustrated in Figure~\ref{fig:errortypes}, these discrepancies can be divided into three categories: \textit{miss}, which stands for the complete omission of certain visual or textual elements (e.g., the top left corner of the shape is missing); \textit{incorrect}, referring to deviations in visual styles or attributes from those specified or expected in the reference slides (e.g., title is not bold); and \textit{disorder}, which describes significant differences in spatial arrangements and alignment of elements compared to the original layout (e.g., the three subheadings are not properly positioned and aligned.).



\textbf{Third, MLLMs' insufficient comprehension of the python-pptx library leads to the generation of syntactically invalid or non-executable code.} Autopresent~\cite{ge2025autopresent} attempts to address this issue by constructing SLIDESLIB, a simplified library built upon python-pptx, encapsulating commonly used operations into a set of high-level APIs. However, this operation inherently restricts the flexibility and comprehensiveness of slide generation. Specifically, SLIDESLIB currently supports only five basic operation types, which neglects more intricate layouts and design requirements commonly encountered in realistic scenarios. Consequently, presentations produced by this approach tend to be overly simplistic, inadequately capturing complex human intentions and detailed visual expectations.



To address the aforementioned limitations, we introduce \scheme, a layout-aware RAG-enhanced hierarchical slide generation framework, which can understand the complex slides and python-pptx library accurately. First, we formulate a novel task, \textbf{Reference Image (RI) to slide generation}, i.e., automatically generating the code for replicating the slide, which is visually consistent with RI. To evaluate the performance of \scheme \ under complex slide scenarios, we propose a novel Slide Complexity Metric (SCM), and construct a new benchmark \bench \ with different difficulty levels based on SCM. Second, we develop a novel \textbf{Color Gradients-based Segmentation} algorithm (\textbf{CGSeg}) that effectively decomposes slide images into semantically meaningful regions. Besides, we propose the \textbf{Layout-aware Prompt}, which integrates the position information of elements to enhance MLLM's understanding of slide layout.  Third, we propose a novel \textbf{Hierarchical Retrieval-Augmented Generation (H-RAG)-based Code Generation} method, which employs a dual-level retrieval-augmented knowledge base~\cite{rag1,rag2} to explicitly enhance MLLMs' understanding of the python-pptx library. At the higher level, a Shape Type Knowledge Base (TS-KB) systematically classifies slide elements and standardizes their descriptions using python-pptx API terminologies. At the lower level, a Operation Function Knowledge Base (OF-KB) captures precise syntactic patterns and invocation paradigms of python-pptx library functions. 

To further enhance the MLLM’s ability to generate high-quality slides, we build a PPTX reverse-engineering tool to construct high quality training data for fine-tuning a 7B model SlideMaster based on Qwen-VL-7B~\cite{qwen}, which can approaches the performance of
the closed-sourced model GPT-4o~\cite{gpt}. Our contributions are summarized as follows: 

\begin{itemize}
    \item We define reference image (RI) to slide generation task and propose a novel Slide Complexity Metric (SCM), based on which we construct \bench, the first difficulty-leveled benchmark with 300 samples.


    \item We propose \scheme, which consists of a novel Color Gradients-based Segmentation algorithm (CGSeg), a Layout-aware Prompt and a Hierarchical Retrieval-Augmented Generation (H-RAG)-based Code Generation method for enhancing the MLLM's understanding on the complex slides and python-pptx library.


    \item We train \model, a 7B open-source model approaching the performance of GPT-4o. To enable effective fine-tuning, we also build a comprehensive PPTX reverse-engineering tool for precise code generation.
\end{itemize}

\section{Related Work}

\subsection{Multimodal Large Language Models for Code Generation}
The multimodal large model demonstrates excellent capabilities in visually rich code generation scenarios, such as UI code generation~\cite{xiao2024interaction2code, xiao2025designbench,ui1,wan2024mrweb}, SVG code generation~\cite{svg1,svg2,svg3,svg4}, and visually rich programming questions~\cite{rich1,rich2,rich3}.
However, MLLMs are not yet capable of plug-and-play use across tasks and still produce subtle errors, therefore, some studies explore their code repair abilities~\cite{repair1,repair2,repair3}.

\subsection{Slide Generation and Understanding}
Previous work on slide generation has predominantly focused on basic content extraction from input documents. With the recent advancements in large language models~\cite{oldppt1,oldppt2,oldppt3,oldppt4}, several studies have begun to explore LLM-based slide generation. For example,~\cite{pptagent} utilizes LLMs to generate slides based on pre-defined slide templates and user-provided text.~\cite{ge2025autopresent} introduces the task of natural language (NL) to slide code generation, aiming to organize visual slide content through textual input. However, its use of coarse-grained natural language descriptions and a native agent design significantly limits the quality of the generated slides.

\section{\bench \ Benchmark}
 
We construct the \bench\ benchmark to evaluate the performance of multimodal large language models (MLLMs) on the Reference Image (RI) to slide generation task. Each instance includes a reference slide image and its corresponding PPTX slide. \bench\ enables comparison of MLLM backbones under varying complexity.~\S\ref{subsec:taskdescription} formally defines the task,~\S\ref{subsec:trim} describes our unified complexity scoring system based on element quantity, diversity, and visual density, and~\S\ref{subsec:datacollection} details data collection and sampling.

\subsection{Task Description}\label{subsec:taskdescription}

This work addresses the task of Reference Image (RI) to slide generation, where the input is a slide's reference image $I_0$ and the goal is to generate Python code using the python-pptx library. Let $F_0$ denote the original slide file corresponding to $I_0$. Given a generation framework $G$ and Multimodal Large Language Models (MLLMs) $M$, the generated code $C_g = G_M(I_0)$ can be executed to obtain a new slide file $F_g$, whose rendered image is denoted as $I_g$. As the original code $C_0$ for $F_0$ is unavailable, we assess the performance of $G$ and $M$ by comparing $(I_0, F_0)$ and $(I_g, F_g)$.

\subsection{Slide Complexity Metric}\label{subsec:trim}
To evaluate slide complexity, we propose a Tri-Metric Slide Complexity Metric (SCM) that integrates production difficulty and visual complexity. Due to the mismatch between visual appearance and construction effort, for example, inserting a visually complex image may require minimal operations. To adress this, we assess slides using: (1) element count, (2) element type count (e.g., textbox, placeholder), and (3) Element Coverage Ratio. The first two reflect operational cost, the third captures visual richness. Since reference complexity labels are not available, we evaluate the relative complexity of sample $i$ within a collection $Y=\{1,2,3,...,N\}$.

Let $c_i$ be the number of elements and $e_i$ the number of distinct element types in sample $i$. The Element Coverage Ratio $v_i$ is the proportion of activated color grids to total grids in the image of sample $i$, computed via the gradient-based segmentation algorithm CGSeg (see~\S\ref{subsec:color} for details).

Each raw dimension score $x_i \in \{c_i, e_i, v_i\}$ is normalized as $\tilde{x}_i = \sigma\left(\frac{x_i - \mu}{\sqrt{\sigma^2} + \epsilon}\right)$, where $\mu$ and $\sigma^2$ denote the mean and variance over all samples in set $Y$, respectively. Here, $\sigma(\cdot)$ is the sigmoid function~\cite{sigmoid}, and $\epsilon$ is a small constant for numerical stability. The final complexity score for slide $i$ is computed via a weighted aggregation: $z_i = \alpha \cdot \tilde{c}_i + \beta \cdot \tilde{e}_i + \gamma \cdot \tilde{v}_i$, where $\alpha + \beta + \gamma = 1$ and the weights $\alpha, \beta, \gamma$ reflect the importance of production effort and visual complexity.
This metric shows a strong correlation with human judgment, as detailed in Section~\S\ref{subsec:tsc_ex}.

\subsection{Data Collection} \label{subsec:datacollection}
To construct a comprehensive benchmark that captures diverse slide characteristics, we randomly sample approximately 32,000 Zenodo10k~\cite{pptagent} slide instances, the largest publicly available slide dataset, to construct the slide set $Y$ as described in~\S\ref{subsec:trim}. To enhance diversity and allow comparative analysis, we additionally incorporate SLIDEBENCH samples in $Y$. This unified set is then used to calculate the normalized complexity scores $z$ for all slides. KMeans algorithm is used to obtain three clusters, whose cluster centers are sorted in order of $z$ to define the simple, medium, and complex levels. From each cluster, we randomly select 100 representative samples from $Y$ to form the final \bench \ benchmark.
\begin{figure}[t]
\small
  \includegraphics[width=\columnwidth]{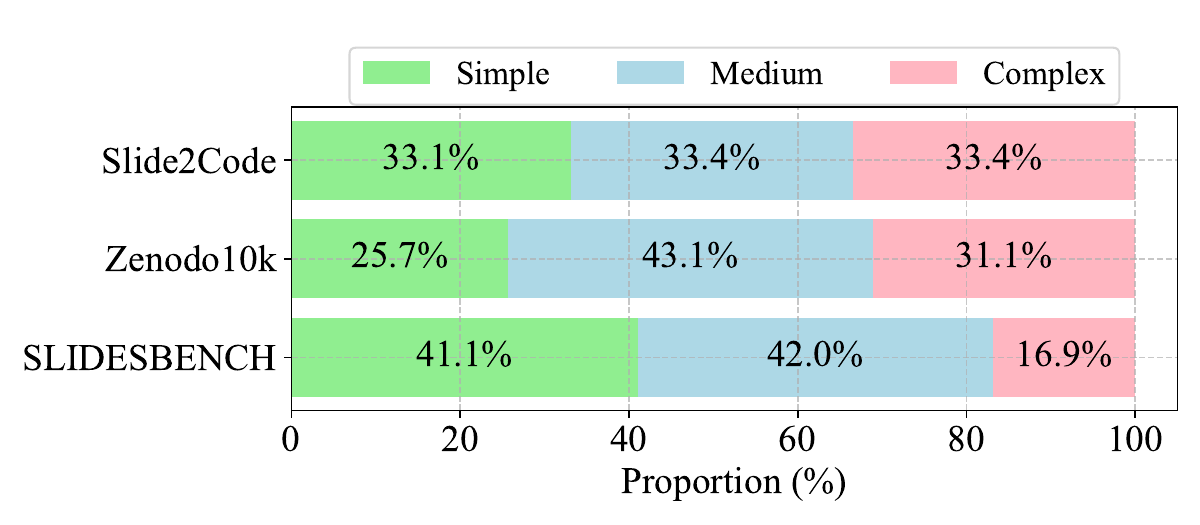}
  \caption{Proportion of samples across three levels in the \bench, Zenodo10k, and SLIDEBENCH datasets.}
  \label{fig:intro_pro}
\end{figure}

Figure~\ref{fig:intro_pro} shows that both Zenodo10k and SLIDEBENCH contain a significantly larger proportion of simple and medium slides. In contrast, \bench \ exhibits a more balanced composition across all three levels, allowing a more equitable evaluation of slide generation models under varying structural and visual complexities.


\section{Methodology}

\newcommand{\segal}{Color Gradients-based Segmentation (CGSeg) algorithm  }
\newcommand{\gener}{Generation}
\newcommand{\hrag}{Hierarchical Retrieval-Augmented Generation(H-RAG)}
\newcommand{\laypmt}{layout-aware prompt}
\newcommand{\one}{\textbf{Describer} }
\newcommand{\two}{\textbf{Coder} }
\newcommand{\three}{\textbf{Assembler} }
\newcommand{\design}{\textit{Design} }
\newcommand{\pics}{\textit{Pictures} }

\begin{figure*}[t]
  \centering
  \includegraphics[width=\textwidth]{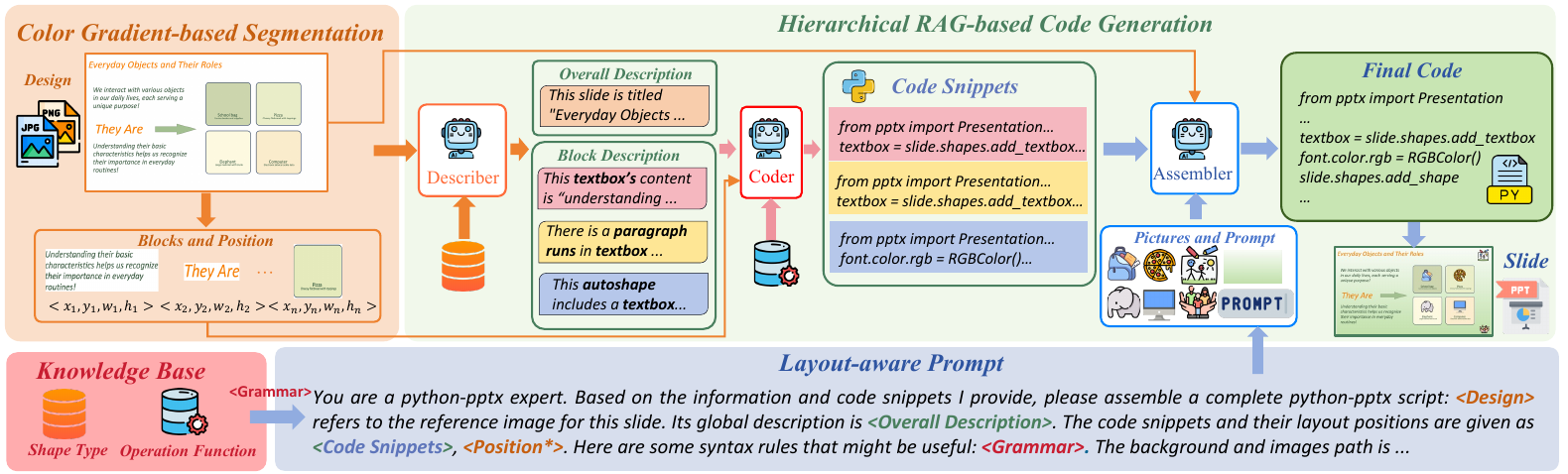}
  \caption{The framework of \scheme.}
  \label{fig:overview}
\end{figure*}

In this section, we introduce \scheme, a unified end-to-end framework for generating Python-executable slide code from reference images (RIs). We assume a scenario where a user provides a design layout ("\design") and embedded visual elements such as pictures or background images ("\pics"). \scheme\ comprises three core modules. First, a \textbf{Color Gradients-based Segmentation} (CGSeg) algorithm segments the input \design\ into semantically meaningful regions. Second, a \textbf{Hierarchical Retrieval-Augmented Code Generation} module, consisting of three collaborative agents \textbf{Describer}, \textbf{Coder}, and \textbf{Assembler}, generates the slide code. Third, a \textbf{Layout-aware Prompt} mechanism enhances the Assembler agent to ensure spatial consistency and syntactic correctness. Finally, based on this framework, we fine-tune a 7B open-source model, named \model.


\subsection{Color Gradient-based Segmentation}\label{subsec:color}

\begin{figure*}[t]
  \centering
  \begin{subfigure}{0.23\textwidth}
    \centering
    \includegraphics[width=\linewidth]{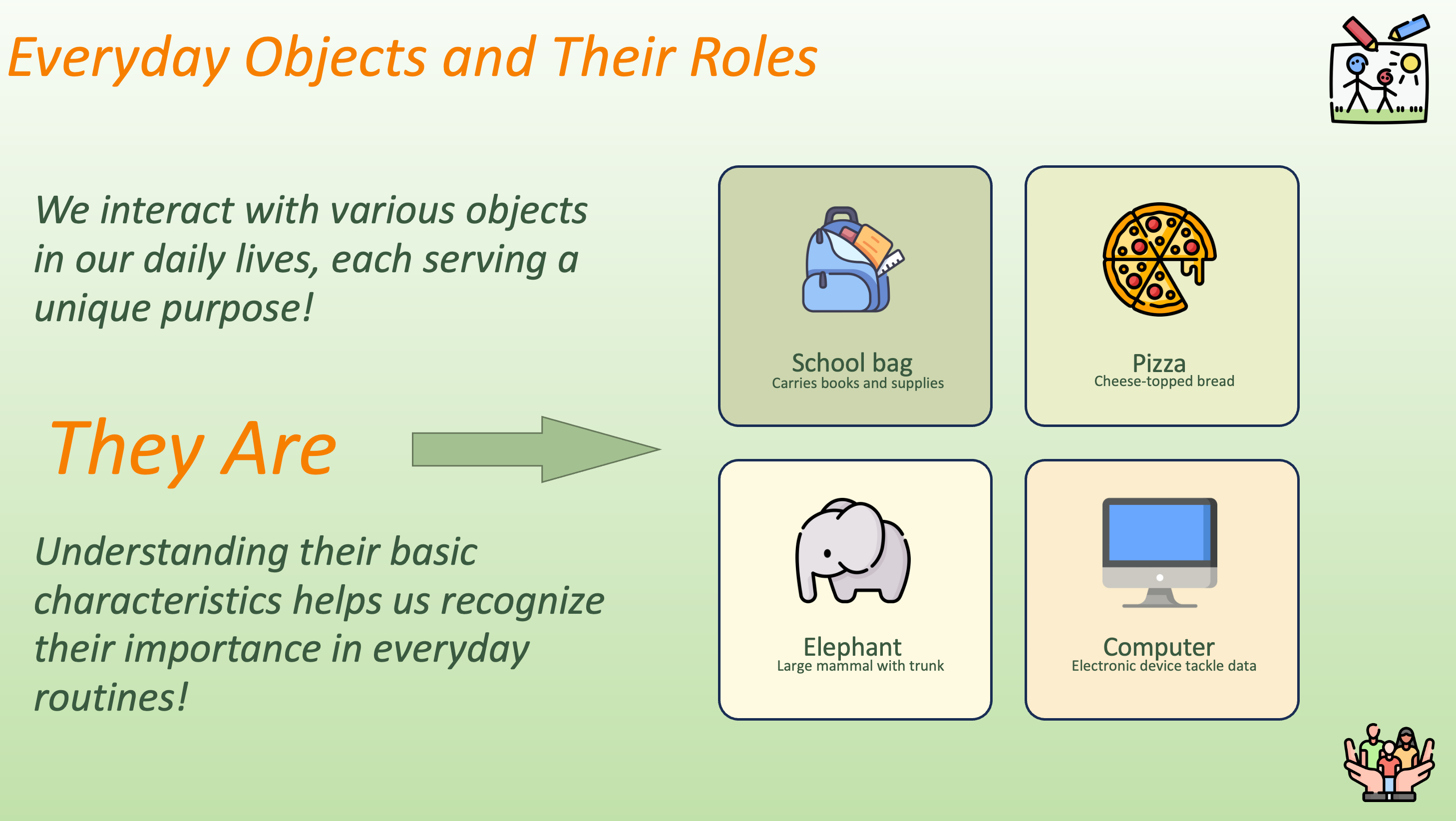}
    \caption{Input Image}
    \label{fig:cgseg-a}
  \end{subfigure}
  \hfill
  \begin{subfigure}{0.23\textwidth}
    \centering
    \includegraphics[width=\linewidth]{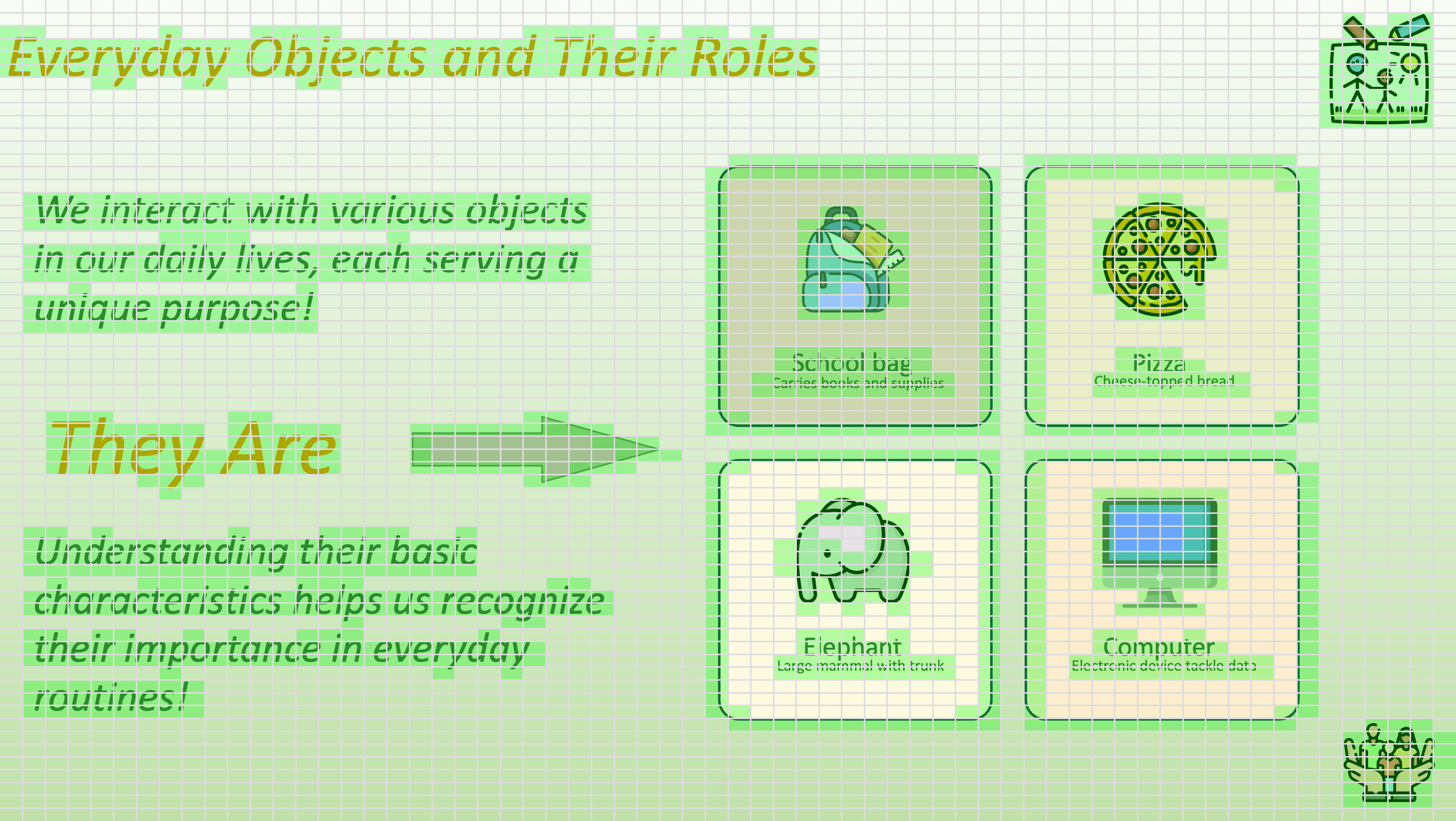}
    \caption{Activated Grid Blocks}
    \label{fig:cgseg-b}
  \end{subfigure}
  \hfill
  \begin{subfigure}{0.23\textwidth}
    \centering
    \includegraphics[width=\linewidth]{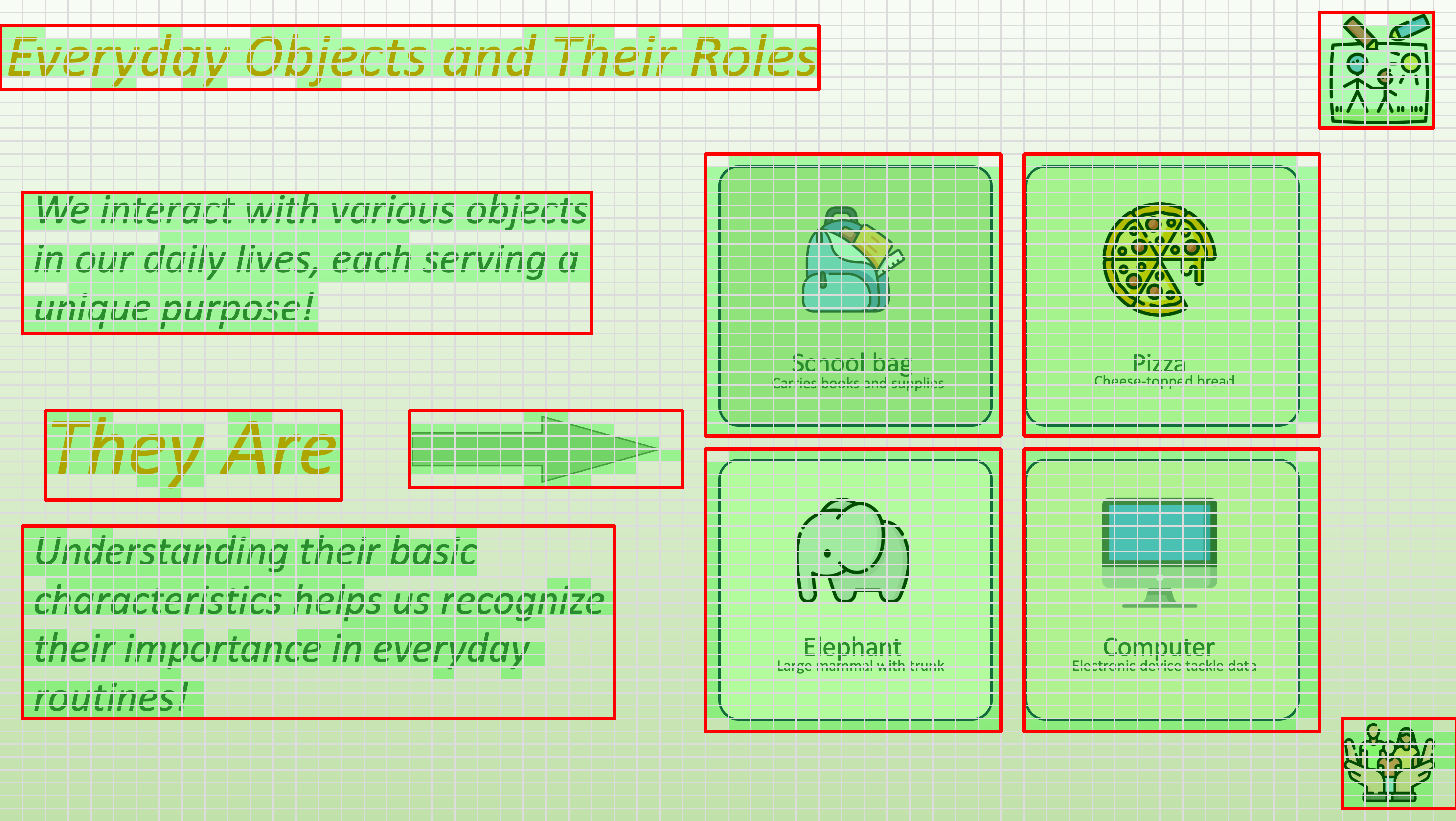}
    \caption{Flood-filled Regions}
    \label{fig:cgseg-c}
  \end{subfigure}
  \hfill
  \begin{subfigure}{0.23\textwidth}
    \centering
    \includegraphics[width=\linewidth]{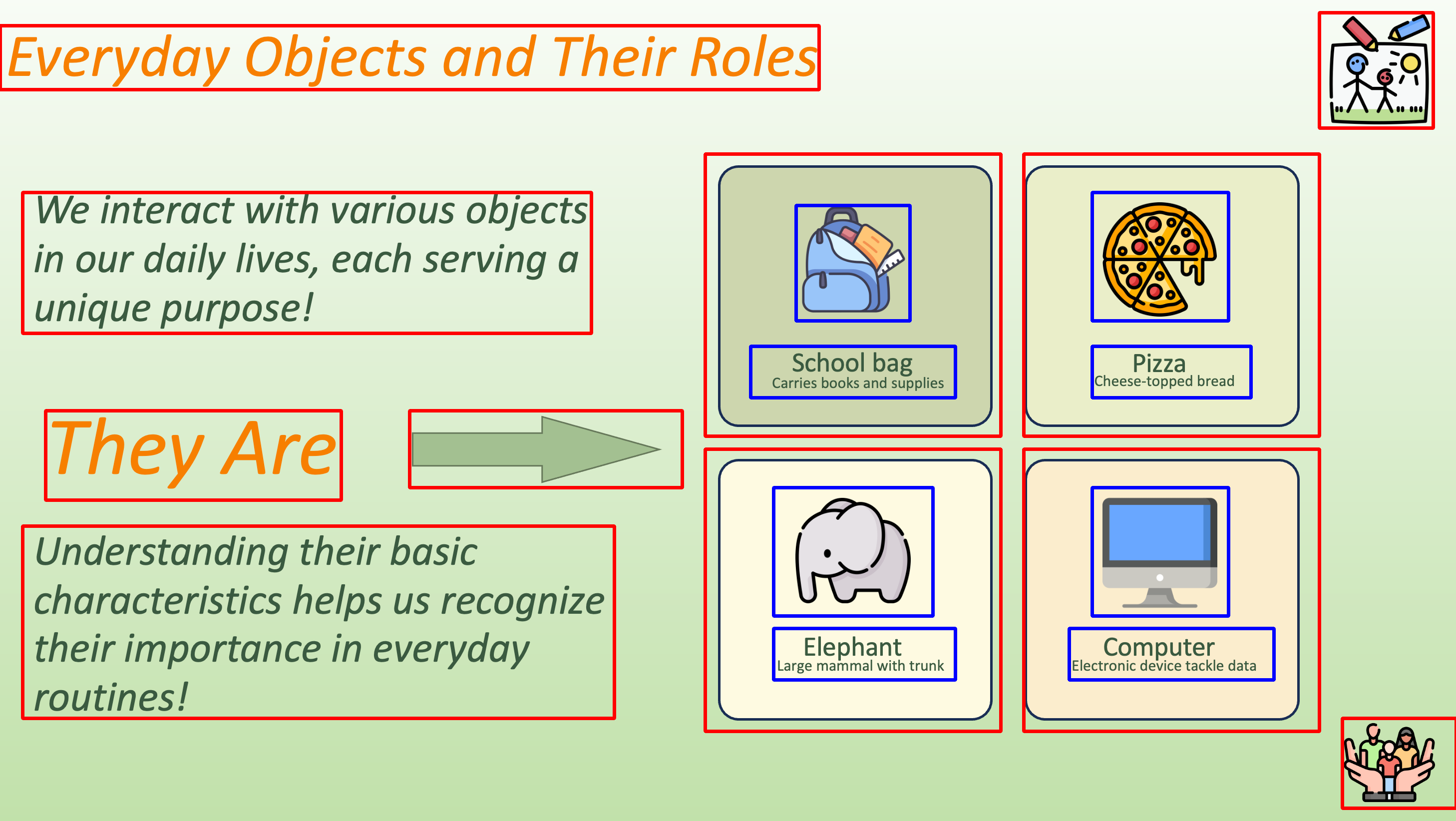}
    \caption{Final result}
    \label{fig:cgseg-d}
  \end{subfigure}

  \caption{An example of CGSeg applied to a slide reference image. The algorithm begins by computing color gradients (a-b), fills them (c), and recursively segments sub-regions (d).}
  \label{fig:cgseg-example}
\end{figure*}

\begin{algorithm}[t]
\small
\caption{Color Gradient-based Segmentation (CGSeg)}
\label{alg:CGSeg}
\begin{algorithmic}[1]
\REQUIRE Image $I$, Grid size $g$, Depth $D$, Max depth $D_{\max}$, Threshold $T$
\ENSURE List of segmented sub-images

\IF{$D = D_{\max}$}
    \RETURN $\emptyset$
\ENDIF

\STATE $G \leftarrow$ \textsc{Split}$(I, g)$ \hfill \texttt{// $g \times g$ grid blocks}
\STATE $C \leftarrow$ \textsc{GradMag}$(G)$ \hfill \texttt{// gradient magnitudes}
\STATE $C_{\text{mid}} \leftarrow$ \textsc{Median}$(C)$ 
\STATE $M \leftarrow \textbf{0}^{g \times g}$ \hfill \texttt{// binary mask}

\FOR{each $c_{ij}$ in $C$}
    \IF{$c_{ij} > T \cdot C_{\text{mid}}$}
        \STATE $M_{ij} \leftarrow 1$ \hfill \texttt{// activate the block}
    \ELSE
        \STATE $M_{ij} \leftarrow 0$
    \ENDIF
\ENDFOR

\STATE $M \leftarrow$ \textsc{Fill}$(M)$ \hfill \texttt{// flood-fill}
\STATE $M_s \leftarrow$ \textsc{Regions}$(M)$ \hfill \texttt{// split connected regions}

\STATE $R \leftarrow \emptyset$ 

\FOR{each $m$ in $M_s$}
    \STATE $I_m,p_m \leftarrow$ \textsc{Crop}$(I, m)$ \hfill \texttt{// get sub-image}
    \STATE add $I_m$ and $p_m$ to $R$
    \STATE $R' \leftarrow$ \textsc{CGSeg}$(I_m, g, D{+}1, D_{\max}, T)$
    \STATE add all in $R'$ to $R$
\ENDFOR

\RETURN $R$
\end{algorithmic}
\end{algorithm}



To reduce the difficulty of MLLM in understanding complex slide design, we proposed CGSeg, a recursive color gradient-based segmentation algorithm to divide slide design into blocks. As shown in Algorithm~\ref{alg:CGSeg}, CGSeg starts by dividing the input image (Figure~\ref{fig:cgseg-a}) into a grid and computing the Sobel magnitude for each block to measure the intensity of the color gradient (lines 4–5). Blocks with gradient magnitudes significantly higher than the median are marked as activated block (lines 6–14), as visualized in Figure~\ref{fig:cgseg-b}. To group visually coherent regions, CGSeg applies a flood-fill~\cite{floodfill} operation to the binary activation mask (line 15), identifying connected regions corresponding to sub-images (line 16), as shown in Figure~\ref{fig:cgseg-c}. These sub-images are further segmented recursively to ensure a hierarchical decomposition of the image $I_m$, along with the corresponding positional information $p_m$ (lines 1–3 and 17–23), with the final segmentation result shown in Figure~\ref{fig:cgseg-d}. This recursive structure allows CGSeg to adaptively refine segment granularity based on local visual complexity, which is crucial for handling slides with heterogeneous layout densities.

\subsection{Hierarchical Retrieval-Augmented Code Generation Module}\label{sub:hragmodule}


\subsubsection{Generation Process}
\label{subsubsec:process}
We design three collaborative MLLM agents whose code generation processes are augmented by H-RAG. 
\one  is responsible for generating a global \design description (Overall Description) as well as block descriptions (Block Description) for each segmented blocks. Based on block and their associated block description, \two produces corresponding code snippets. Subsequently, \three generates the complete slide code by \laypmt, which will be elaborated in~\S\ref{subsec:prompt}, along with the \pics provided. Executing this code produces a slide that structurally and visually aligns with the Reference Image(RI). If the generated code is not executable \three applies a self-refinement mechanism to correct syntax errors, where errors serves as the feedback to prompt the MLLM to re-generate the code.


Beyond the above inputs, each agent draws knowledge from distinct bases according to its role. The form and origin of the knowledge used in each agent’s prompt are detailed in~\S\ref{subsubsec:hrag}.

\subsubsection{Hierarchical Retrieval-Augmented Generation} \label{subsubsec:hrag}

\hrag\ comprises a Shape Type Knowledge Base and an Operation Function Knowledge Base. The former contains descriptions of objects from the python-pptx documentation, used in \one\ to guide standardized description generation. For example, in “This \textbf{\textit{autoshape}} includes a \textbf{\textit{textbox}}...”, both terms are object names from the documentation. The latter includes full syntax specifications (e.g., parameters, return values, etc.). Appendix~\ref{sub-app:kb} details their structure.

We employ BGE M3-Embedding~\cite{bme} to embed entries and build a vector-based retrieval database. For a prompt $p$, its vector $q_p$ is computed, and cosine similarity $\cos(q_p, k_i)$ is used to match $k_i$. The top-$k$ relevant entries are inserted into $p$. Given the size of the Shape Type Knowledge Base, all entries are included in \one\ to ensure complete type coverage.

In the hierarchical pipeline, agents collaborate progressively. \one retrieves object types from the Shape Type Knowledge Base to identify elements in block images and output standardized descriptions. \two uses these to query the Operation Function Knowledge Base and generate code snippets. \three uses these snippets to retrieve full syntax patterns and generate executable code.


\subsection{Layout-aware Prompt} \label{subsec:prompt}
\definecolor{Designc}{HTML}{bf5311}
\definecolor{Descric}{HTML}{3f7353}
\definecolor{Codec}{HTML}{5369b1}
\definecolor{Grammarc}{HTML}{c01b2b}
After \two completes the generation of code snippets for blocks, \three  is applied to assemble these code snippets for generating the final slide in an accurate manner. The assembly prompt needs to meet the following requirements: (1) ensure that each block appears in the correct position in the final slide; (2) avoid syntax errors in the merged code and ensure code context consistency.

To achieve above goals, layout-aware prompt injects the layout position using python-pptx standard positioning units (inches) to ensure the position correctness and retrieve the grammar \textcolor{Grammarc}{\textbf{<Grammar>}} from Knowledge Base to avoid syntax errors and code conflicts. Since the resolution of the \design differs from the actual slide layout size, we apply proportional scaling to the Position (<$x, y, w, h$>) extracted from \segal to map it onto the slide coordinates, denoted as \textcolor{Designc}{\textbf{<Position*>}}. Subsequently, the reference image design \textcolor{Designc}{\textbf{<Design>}}, global body description \textcolor{Descric}{\textbf{<Overall description.>}}, partial codes \textcolor{Codec}{\textbf{<Code Snippets>}} from \textbf{Coder}, layout representation \textcolor{Designc}{\textbf{<Position*>}}, and syntactic patterns \textcolor{Grammarc}{\textbf{<Grammar>}} retrieved from the \hrag\ knowledge base are integrated into a predefined prompt template to construct the final layout-aware prompt (see Appendix~\ref{sub-app:prompt} for details).

\subsection{\model} \label{subsec:model}
Using the SLIDESBENCH training set, we construct a dataset of (RI, instruction, program) triplets. The reverse-engineering tool proposed by~\cite{ge2025autopresent} produces labels (Python code) for only a limited set of slide styles, resulting in suboptimal training data quality. To mitigate this, we develop a new reverse-engineering tool capable of handling a broader spectrum of slide styles, thereby enhancing label quality. The effectiveness of this tool is analyzed in~\S\ref{subsec:rever}. We fine-tune our model, SlideMaster, based on Qwen2.5-VL-7B-Instruct~\cite{qwen}, using LoRA~\cite{lora}. Full configuration details are provided in Appendix~\ref{sec-app:lora}.

\section{Experiments and Results}


\subsection{Experimental Setup}\label{subsec:set}

\textbf{Model}. To evaluate the performance of the SlideCoder, we employ state-of-the-art (SOTA) models, including GPT-4o~\cite{gpt}, Gemini-2.0-flash~\cite{gemini}, and SlideMaster, which is a fine-tuned model based on the open-source Qwen2.5-VL-7B-Instruct~\cite{qwen}. The SOTA models are accessed via their official APIs, with GPT-4o using version 20241120 and Gemini-2.0-flash accessed in May 2025. For both models, the maximum token limit and temperature are set to 4096 and 0, respectively. Same as~\cite{ge2025autopresent}, we allow both \two and \three agents up to three self-refinement attempt. The first successful attempt is taken as the output. If \two fails to generate executable code after the maximum number of attempts, the corresponding block is discarded. If \three fails, the corresponding sample is marked as execution failure.

\textbf{Metric}. To comprehensively assess generation quality, we adopt four metrics, using the notations defined in~\S\ref{subsec:taskdescription}. (1) \textbf{Global Visual Metrics}, including CLIP~\cite{clipscore} and SSIM~\cite{ssim} scores computed between the original image $I_0$ and the generated image $I_g$; (2) \textbf{Local Structural Metrics}, which compare the original and generated slide files $F_0$ and $F_g$ in terms of content similarity and position similarity, following~\cite{ge2025autopresent}; (3) \textbf{Execution}, defined as the success rate of executing $C_g$ without errors; and (4) \textbf{Overall Score}, calculated as the average of all metric values across all samples, with failed executions assigned a score of zero.
\subsection{Quantitative Results and Analysis}\label{subsec:mainexp}

\definecolor{tg}{RGB}{100, 141, 63}
\definecolor{ty}{RGB}{176, 143, 46}
\definecolor{tr}{RGB}{123, 32, 38}
\newcommand{\textg}[1]{\textbf{\textcolor{tg}{#1}}}  
\newcommand{\texty}[1]{\textbf{\textcolor{ty}{#1}}}  
\newcommand{\textr}[1]{\textbf{\textcolor{tr}{#1}}}  

\begin{table*}[h!]
\small
    \centering
    \caption{Results on Slide2Code (top) and SLIDESBENCH (bottom) using SlideCoder and AutoPresent with different MLLMs. \colorbox{color11}{Green}, \colorbox{color2}{yellow}, and \colorbox{low}{red} indicate simple, medium, and complex levels in SlideCoder. \textbf{Bolded values} mark the best result per level.}
    \label{tab:main_exper}
    \begin{tabular}{c|c|c|cc|cc|c}
    \toprule
       \multirow{2}{*}{\bf Framework}&\multirow{2}{*}{\bf Backbone}&\multirow{2}{*}{\bf Execution\%}&\multicolumn{2}{c|}{\bf Local Structural Metrics} &\multicolumn{2}{c|}{\bf Global Visual Metrics}&\multirow{2}{*}{\bf Overall}\\\cline{4-7}
&&&\makebox[1.6cm][c]{\textbf{Content}} &
\makebox[1.6cm][c]{\textbf{Position}} &
\makebox[1.6cm][c]{\textbf{Clip}} &
\makebox[1.6cm][c]{\textbf{SSIM}} &\\\hline
\multicolumn{8}{c}{\cellcolor{color1}\it Slide2Code}\\\hline
\multirow{9}{*}{AutoPresent}&\multirow{3}{*}{AutoPresent}&\cellcolor{color11}61.0 &\cellcolor{color11}92.7 &\cellcolor{color11}78.9 &\cellcolor{color11}70.8 &\cellcolor{color11}80.3 &\cellcolor{color11}48.6 \\
&&\cellcolor{color2}53.0 &\cellcolor{color2}89.6 &\cellcolor{color2}77.3 &\cellcolor{color2}69.2 &\cellcolor{color2}79.1 &\cellcolor{color2}41.4 \\
&&\cellcolor{low}67.0 &\cellcolor{low}87.2 &\cellcolor{low}71.4 &\cellcolor{low}65.9 &\cellcolor{low}73.4 &\cellcolor{low}48.5\\ \cline{3-8}
&\multirow{3}{*}{Gemini2.0-flash}&\cellcolor{color11}57.0 &\cellcolor{color11}91.4 &\cellcolor{color11}78.3 &\cellcolor{color11}69.7 &\cellcolor{color11}79.0 &\cellcolor{color11}44.8 \\
&&\cellcolor{color2}68.0 &\cellcolor{color2}88.7 &\cellcolor{color2}79.9 &\cellcolor{color2}66.3 &\cellcolor{color2}71.6 &\cellcolor{color2}51.5 \\
&&\cellcolor{low}66.0 &\cellcolor{low}89.3 &\cellcolor{low}72.2 &\cellcolor{low}63.1 &\cellcolor{low}64.7 &\cellcolor{low}45.2 \\\cline{3-8}
&\multirow{3}{*}{GPT-4o}&\cellcolor{color11}58.0 &\cellcolor{color11}92.7 &\cellcolor{color11}80.9 &\cellcolor{color11}68.8 &\cellcolor{color11}75.6 &\cellcolor{color11}45.4 \\
&&\cellcolor{color2}50.0 &\cellcolor{color2}92.3 &\cellcolor{color2}74.6 &\cellcolor{color2}67.6 &\cellcolor{color2}72.6 &\cellcolor{color2}36.8 \\
&&\cellcolor{low}69.0 &\cellcolor{low}90.3 &\cellcolor{low}73.3 &\cellcolor{low}62.3 &\cellcolor{low}63.3 &\cellcolor{low}47.1 \\\hline
\multirow{9}{*}{SlideCoder}&\multirow{3}{*}{SlideMaster}&\cellcolor{color11}86.0 &\cellcolor{color11}92.4 &\cellcolor{color11}87.4 &\cellcolor{color11}77.6 &\cellcolor{color11}91.1 &\cellcolor{color11}76.7 \\
&&\cellcolor{color2}75.0 &\cellcolor{color2}84.7 &\cellcolor{color2}79.8 &\cellcolor{color2}75.4 &\cellcolor{color2}\texty{86.4} &\cellcolor{color2}61.7 \\
&&\cellcolor{low}73.0 &\cellcolor{low}76.1 &\cellcolor{low}70.5 &\cellcolor{low}72.4 &\cellcolor{low}{\textr{82.8}} &\cellcolor{low}54.2 \\\cline{3-8}
&\multirow{3}{*}{Gemini2.0-flash}&\cellcolor{color11}97.0 &\cellcolor{color11}94.5 &\cellcolor{color11}\textg{88.6} &\cellcolor{color11}\textg{81.3} &\cellcolor{color11}90.7 &\cellcolor{color11}87.0 \\
&&\cellcolor{color2}90.0 &\cellcolor{color2}90.9 &\cellcolor{color2}84.6 &\cellcolor{color2}\texty{82.3} &\cellcolor{color2}85.5 &\cellcolor{color2}76.6 \\
&&\cellcolor{low}88.0 &\cellcolor{low}92.7 &\cellcolor{low}{\textr{80.9}} &\cellcolor{low}{\textr{81.7}} &\cellcolor{low}81.2 &\cellcolor{low}71.6 \\\cline{3-8}
&\multirow{3}{*}{GPT-4o}&\cellcolor{color11}\textg{99.0} &\cellcolor{color11}\textg{96.3} &\cellcolor{color11}88.1 &\cellcolor{color11}79.8 &\cellcolor{color11}\textg{91.8} &\cellcolor{color11}\textg{89.1} \\
&&\cellcolor{color2}\texty{100.0} &\cellcolor{color2}\texty{92.5} &\cellcolor{color2}\texty{84.7} &\cellcolor{color2}81.5 &\cellcolor{color2}86.2 &\cellcolor{color2}\texty{85.5} \\
&&\cellcolor{low}{\textr{96.0}} &\cellcolor{low}{\textr{94.3}} &\cellcolor{low}80.0 &\cellcolor{low}80.7 &\cellcolor{low}82.6 &\cellcolor{low}{\textr{78.4}}\\\hline
\multicolumn{8}{c}{\cellcolor{color1}\it SLIDESBENCH}\\\hline
\multirow{3}{*}{AutoPresent}&AutoPresent&84.1 &92.2 &67.2 &81.6 &73.7 &65.3 \\
&Gemini2.0-flash&56.4 &91.7 &62.9 &77.1 &66.0 &40.4\\ 
&GPT-4o&86.7 &92.5 &76.3 &78.0 &70.8 &66.9 \\\hline
\multirow{3}{*}{SlideCoder}&SlideMaster&87.2 &91.5 &76.9 &73.4 &80.0 &68.4 \\
&Gemini2.0-flash&89.7 &90.0 &{\bf 85.4} &{81.8} &80.0 &75.0 \\
&GPT-4o&{\bf 94.9} &{\bf 94.8} &83.9 &\bf{82.1} &\bf{80.9} &{\bf 78.8} \\\bottomrule
    \end{tabular}
\end{table*}

The upper part of Table~\ref{tab:main_exper} presents the performance of different frameworks on our proposed benchmark, evaluated using the metrics introduced in Section~\ref{subsec:taskdescription}. The results show that SlideCoder consistently achieves the best performance across all difficulty levels. Specifically, its overall score surpasses the best baseline by 40.5, 34.0, and 29.9 points on the simple, medium, and complex levels, respectively, demonstrating the overall superiority of our framework. For execution success rate, SlideCoder outperforms the best baseline by 38\%, 32\%, and 27\% across the three difficulty levels, indicating that the proposed H-RAG and CGSeg mechanisms significantly enhance model performance and reduce task difficulty.

Moreover, \scheme\  outperforms all baselines in both Local Structural Metrics and Global Visual Metrics, confirming its strong fidelity in preserving both the structural layout and visual appearance of the original slides. The stepwise decline in \scheme’s overall score across increasing difficulty levels further indicates its ability to leverage visual and structural cues from the input slides. In contrast, baseline models relying solely on natural language descriptions exhibit weak sensitivity to slide complexity, failing to reflect the difficulty hierarchy in their overall scores.

On the SLIDESBENCH dataset (as shown in the lower part of Table~\ref{tab:main_exper}), \scheme\  also surpasses all baselines across all metrics, with an overall score of 78.8 when using GPT-4o as the backbone, representing a 11.9 improvement over the best-performing baseline. Notably, the open-source fine-tuned model \model\  also demonstrates competitive performance, outperforming the best GPT-4o-based baseline on both datasets.

\subsection{Reverse Tool Analysis}\label{subsec:rever}

\begin{table}[ht]
\centering
\small
\caption{Object Types and Corresponding Style count}
\begin{tabular}{l!{\vrule width 0.2pt}cc}
    \toprule
    \textbf{Type Name} & \textbf{Ours} & \textbf{AutoPresent's} \\
    \midrule
    \texttt{title} & 10 & 3 \\
    \texttt{textbox} & 10 & 5 \\
    \texttt{bullet points} & 8 & 5 \\
    \texttt{background color} & 1 & 1 \\
    \texttt{image} & 2 & 2 \\
    \texttt{placeholder} & 4 & -- \\
    \texttt{freeform} & 2 & -- \\
    \texttt{connector} & 5 & -- \\
    \texttt{table} & 4 & -- \\
    \texttt{triangle} & 5 & -- \\
    \bottomrule
\end{tabular}
\label{tab:slide-attributes}
\end{table}

Table~\ref{tab:slide-attributes} summarizes the supported object types and corresponding styles in our proposed reverse engineering tool. Our tool supports 10 commonly used object types and 44 distinct object styles, whereas Autopresent~\cite{ge2025autopresent} only supports 5 object types and 16 styles. Detailed comparisons can be found in Appendix~\ref{sub-app:rever}. To quantitatively assess the reverse engineering capabilities of both tools, we adopt two evaluation metrics:

\textbf{Reconstruction Ratio}: This metric calculates the ratio between the number of shapes in the slide reconstructed from the reverse-engineered code and the original slide. Our tool achieves a reconstruction ratio of 90.38\%, significantly outperforming~\cite{ge2025autopresent}, which only reaches 65.67\%. This demonstrates the broader object type coverage enabled by our tool.

\textbf{CLIP Score}: Our method achieves a CLIP score~\cite{clipscore} of 88.66\%, whereas Autopresent~\cite{ge2025autopresent} only achieves 69.87\%. The higher score indicates that our reverse-engineered slides more accurately preserve the visual and stylistic details of the original, owing to the broader support for object types and styles.


\subsection{Slide Complexity Metric Analysis}\label{subsec:tsc_ex}
\begin{figure*}[ht]
  \centering
  \small
  \includegraphics[width=\textwidth]{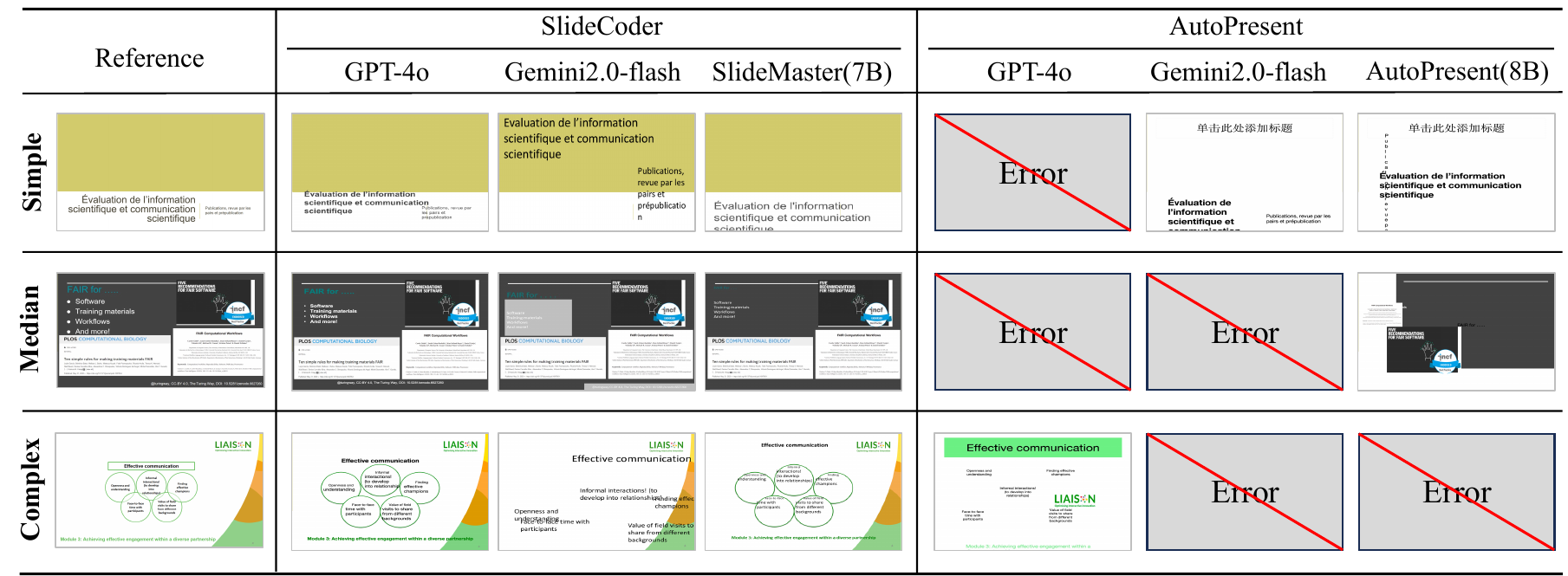}
  \caption{Examples of slides generated by different methods in three difficulty levels.}
  \label{fig:case}
\end{figure*}

To evaluate the effectiveness of the proposed  Slide Complexity Metric (SCM), we conducted a human subject study. A total of 100 samples were randomly selected from the Slide2Code benchmark for evaluation. Four doctoral students were recruited as annotators, each assigned 50 slides to assess. The annotators were instructed to score each slide from the perspective of three dimensions: the number of shapes, the diversity of shape types, and the level of element coverage. The scoring range was 0–100, following the protocol in Appendix~\ref{sub-app:guideline}. Each slide was rated independently by two annotators, and the final score was their average.

To assess the alignment between SCM and human perception, we first compute the Pearson correlation coefficient~\cite{pearson} between the SCM complexity scores and the averaged human scores. The result is $r = 0.873$ with a p-value of $2.776 \times 10^{-32}$, indicating a strong and statistically significant correlation. Additionally, we calculated the intraclass correlation coefficient~\cite{icc} between the SCM scores and each individual annotator’s score to assess consistency. The ICC result is $0.726$ with a p-value of $1.186 \times 10^{-31}$, demonstrating substantial agreement between SCM and human evaluations. These results confirm that SCM is a reliable and objective metric aligned with human judgment of slide complexity.

\subsection{Ablation Study}\label{subsec:ablation}

\begin{table}[h!]
    \centering
    \small
    \caption{Overall performance of ablation study. }
    \label{tab:overall_ablation}
    \begin{tabular}{c|c|c}
    \toprule
\bf{Setting}&\bf{Execution\%}&\bf{Overall}\\\hline
\multirow{3}{*}{ SlideCoder}&\cellcolor{color11}\textg{100.0} &\cellcolor{color11}\textg{89.9}\\ 
&\cellcolor{color2}\texty{100.0} &\cellcolor{color2}\texty{85.8} \\
&\cellcolor{low}\textr{100.0} &\cellcolor{low}\textr{82.2} \\\cline{1-3}
\multirow{3}{*}{w/o Layout}&\cellcolor{color11}100.0 &\cellcolor{color11}81.2\\ 
&\cellcolor{color2}93.9 &\cellcolor{color2}73.6\\ 
&\cellcolor{low}93.9 &\cellcolor{low}71.8  \\\cline{1-3}
\multirow{3}{*}{w/o CGSeg}&\cellcolor{color11}75.8 &\cellcolor{color11}55.4\\ 
&\cellcolor{color2}51.5 &\cellcolor{color2}39.6 \\
&\cellcolor{low}69.7 &\cellcolor{low}48.4 \\\cline{1-3} 
\multirow{3}{*}{w/o H-RAG}&\cellcolor{color11}90.9 &\cellcolor{color11}80.4 \\
&\cellcolor{color2}81.8 &\cellcolor{color2}69.3\\ 
&\cellcolor{low}84.8 &\cellcolor{low}70.7
\\\cline{1-3} 
\multirow{3}{*}{Native Setting} &\cellcolor{color11}75.8 &\cellcolor{color11}53.9 \\
&\cellcolor{color2}48.5 &\cellcolor{color2}37.4\\ 
&\cellcolor{low}66.7 &\cellcolor{low}46.9
\\
\bottomrule
\end{tabular}
\end{table}

We design three ablation settings to validate the effectiveness of different components in our framework: (1) w/o Layout, removes the layout-aware prompt; (2) w/o CGSeg, disables both the CGSeg mechanism and the layout-aware prompt; (3) w/o H-RAG, removes the \textcolor{Grammarc}{\textbf{<Grammar>}} content from all prompts.(4) Native setting, which removes H-RAG on top of the w/o CGSeg setting. Detailed descriptions are provided in Appendix~\ref{app-sub:AblSetting}. We randomly sample 33 instances from each difficulty level, resulting in a total of 99 samples, and perform inference using GPT-4o. The overall results are reported in Table~\ref{tab:overall_ablation}, with detailed metric result provided in Appendix~\ref{app-subsec:ablres}. After removing each component, both execution rate and overall score exhibit varying degrees of decline, which demonstrates the contribution of each component to the overall framework. Notably, the w/o CGSeg setting shows significant performance drops across all metrics. Although slightly better than the Native setting due to the presence of H-RAG.

\subsection{Case Study}\label{subsec:case}

Figure~\ref{fig:case} presents slides generated by different models under three levels of difficulty. It can be observed that models based on natural language often fail to satisfy the detailed and layout-specific requirements of reference images. These models frequently produce slides with overlapping elements or content that extends beyond canvas boundaries. In medium and complex samples, the generated code often fails to compile. In contrast, SlideCoder’s CGSeg mechanism enables the MLLM to focus more effectively on fine-grained details. Moreover, the layout-aware prompt helps ensure that the spatial arrangement of elements aligns more closely with reference image.

\section{Conclusion}
We introduce a new Reference Image to Slide Generation task and a novel Slide Complexity Metric for evaluating slide complexity. Based on this metric, we build the \bench \ benchmark with different levels of difficulty. We also propose \scheme \ enhanced by a Color Gradients-based Segmentation algorithm, a Layout-aware Prompt and a Hierarchical Retrieval-Augmented Code Generation for accurate slide generation. A high-quality training set is curated to fine-tune a 7B open-source model. Experimental results show that \scheme \ outperforms the strongest baselines.

\section*{Limitations}

In this work, we take the first step toward vision-based slide generation. While our method achieves substantial improvements across multiple evaluation metrics, several limitations remain unaddressed. First, the current framework focuses on generating a single slide from one reference image and does not explore the multi-slide generation scenario. Second, we assume that user input contains separate design and image components, and do not handle the case where a complete slide with embedded pictures is provided as input. Third, due to budget and time constraints, our segmentation algorithm adopts a fixed-rule paradigm. Future work may investigate more flexible model-based detection approaches to enable adaptive and accurate block partitioning.

\bibliography{custom}

\appendix
\section{Detail ablation analysis}
\label{sub-app:ablation}
\subsection{Details of Ablation Settings}
\label{app-sub:AblSetting}
\begin{itemize}
    \item w/o Layout: Removes only the layout-aware prompt, meaning that the input to \three\ does not contain the positional coordinates of each block.
    \item w/o CGSeg: Disables the CGSeg mechanism. Since the goal of \two\ is to generate partial code and \three\ is responsible for code assembly, the removal of CGSeg renders \three\ unnecessary. Consequently, both \three\ and its layout-aware prompt are removed in this setting, and the output code generated by \two\ is directly treated as the final output of the framework.
    \item w/o H-RAG: Disables the retrieval of knowledge base content for all agents.
    \item Native setting: Disables both H-RAG and CSeg components. Specifically, we input ordinary prompts that do not incorporate H-RAG, allowing the MLLMs to generate complete slide code directly from the reference image. This setup is used to evaluate the baseline capability of native MLLMs in handling the reference image to slide code generation task.

\end{itemize}

\subsection{Detailed Analysis of Ablation Results}
\label{app-subsec:ablres}
Table~\ref{tab:detail_ablation} provides a detailed evaluation metrics under different ablation settings. 

\textbf{In the w/o Layout setting}, the Position score under the complex level drops significantly from 81.35 to 72.16. This is primarily because, in complex cases, the CGSeg algorithm typically divides the Reference Image(RI) into more blocks, and without layout information, the Agent struggles to model spatial relationships among multiple elements. This often leads to overlapping or out-of-bound content, causing a sharp decline in the Position metric and slightly affecting other metrics as well.

\textbf{In the w/o CGSeg setting}, both the CGSeg mechanism and the layout-aware prompt are removed. As a result, a single \one\ Agent is required to handle the entire complex slide, which exceeds its processing capacity, often leading to code generation failures and a sharp drop in execution success rate. Its performance is slightly better than the Native setting due to the additional knowledge provided by H-RAG.

\textbf{In the w/o H-RAG setting}, the \textcolor{Grammarc}{\textbf{<Grammar>}} component is removed from each Agent. Excluding this component from \one\ reduces its ability to accurately identify the corresponding python-pptx object. Similarly, removing it from \two\ and \three\ deprives the Agents of essential syntactic guidance, often resulting in version-related errors caused by inconsistencies between the model's training data and the current version of the python-pptx library. These combined factors lead to overall performance degradation.

\textbf{In the Native setting}, both the CGSeg mechanism and H-RAG are removed, leaving a single \two\ Agent to handle the entire slide without any auxiliary support. This reduces the framework to a plain MLLM-based inference process, severely limiting its ability to generate structured and executable code, and resulting in the lowest execution rate and overall performance.

\section{Detailed comparisons of Reverse Tool}
\label{sub-app:rever}

Table~\ref{tab:tool-our} lists the object types and their styles supported by our reverse engineering tool. 

Table~\ref{tab:tool-ap} lists the object types and their styles supported by AutoPresent's reverse engineering tool. 

\section{LoRA fine-tuning parameters}
\label{sec-app:lora}
The LoRA fine-tuning parameters are listed in Table~\ref{tab:lora-config}.

\section{Evaluation Dimensions and Scoring Criteria}
\label{sub-app:guideline}
\begin{figure*}[ht]
  \centering
  \includegraphics[width=\textwidth]{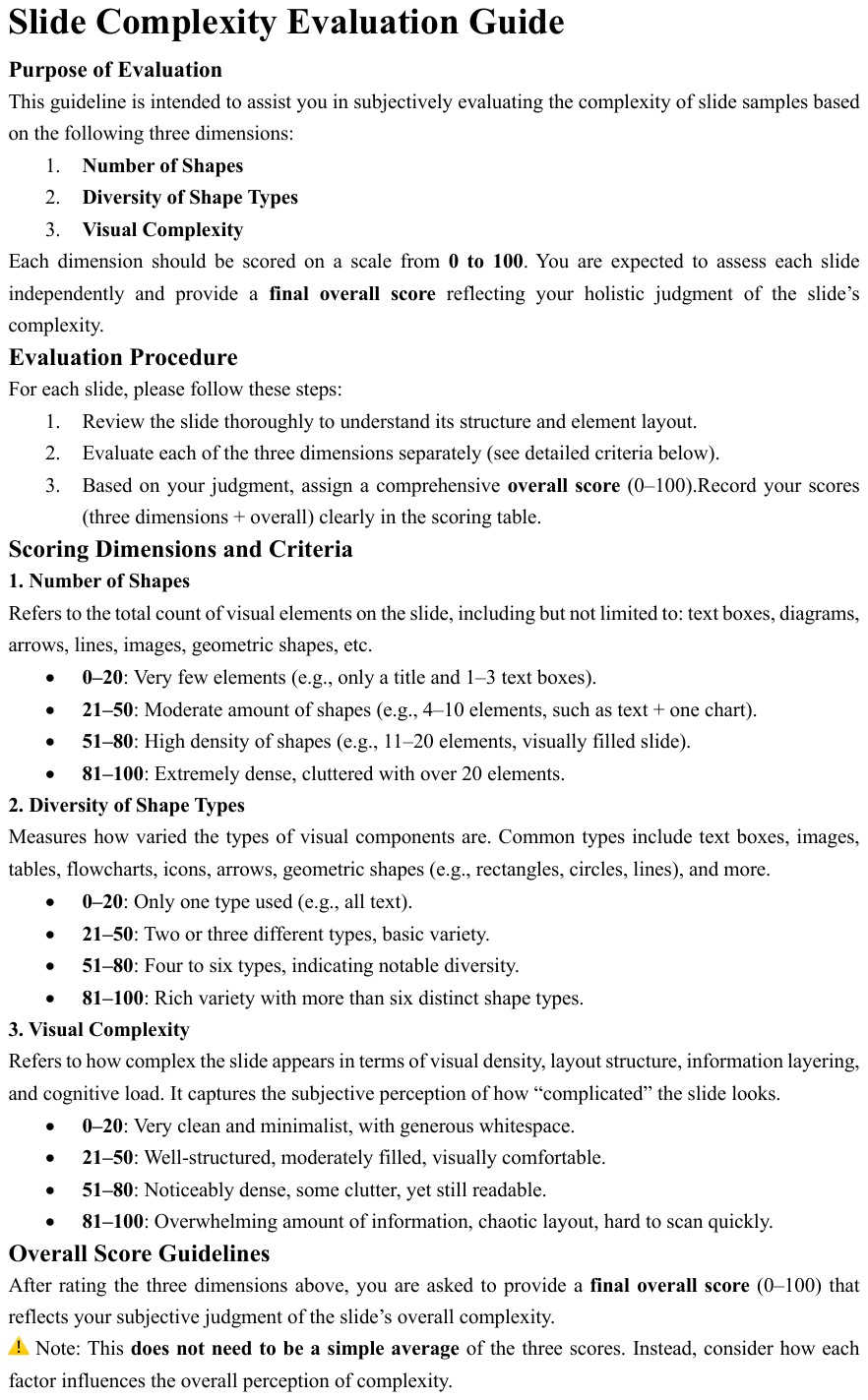 }
  \caption{Evaluation guidelines provided to the four doctoral student annotators.}
  \label{fig:guidelines}
\end{figure*}

The evaluation guidelines for the four doctoral student annotators are provided in Figure~\ref{fig:guidelines}.

\section{Prompt Templates}
\label{sub-app:prompt}
\begin{figure*}[ht]
  \centering
  \includegraphics[width=\textwidth]{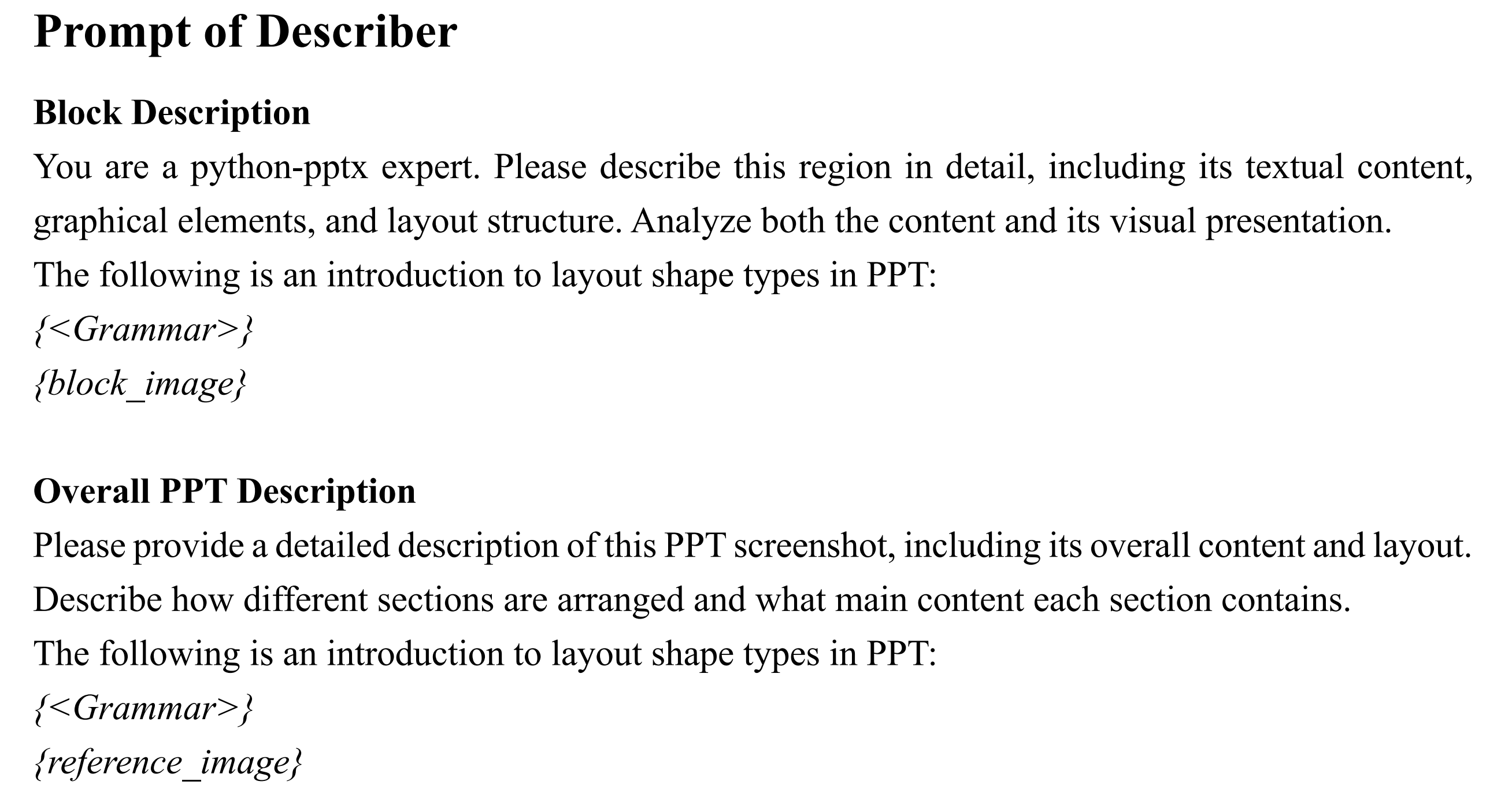}
  \caption{Prompt of Describer.}
  \label{fig:prompt1}
\end{figure*}

\begin{figure*}[ht]
  \centering
  \includegraphics[width=\textwidth]{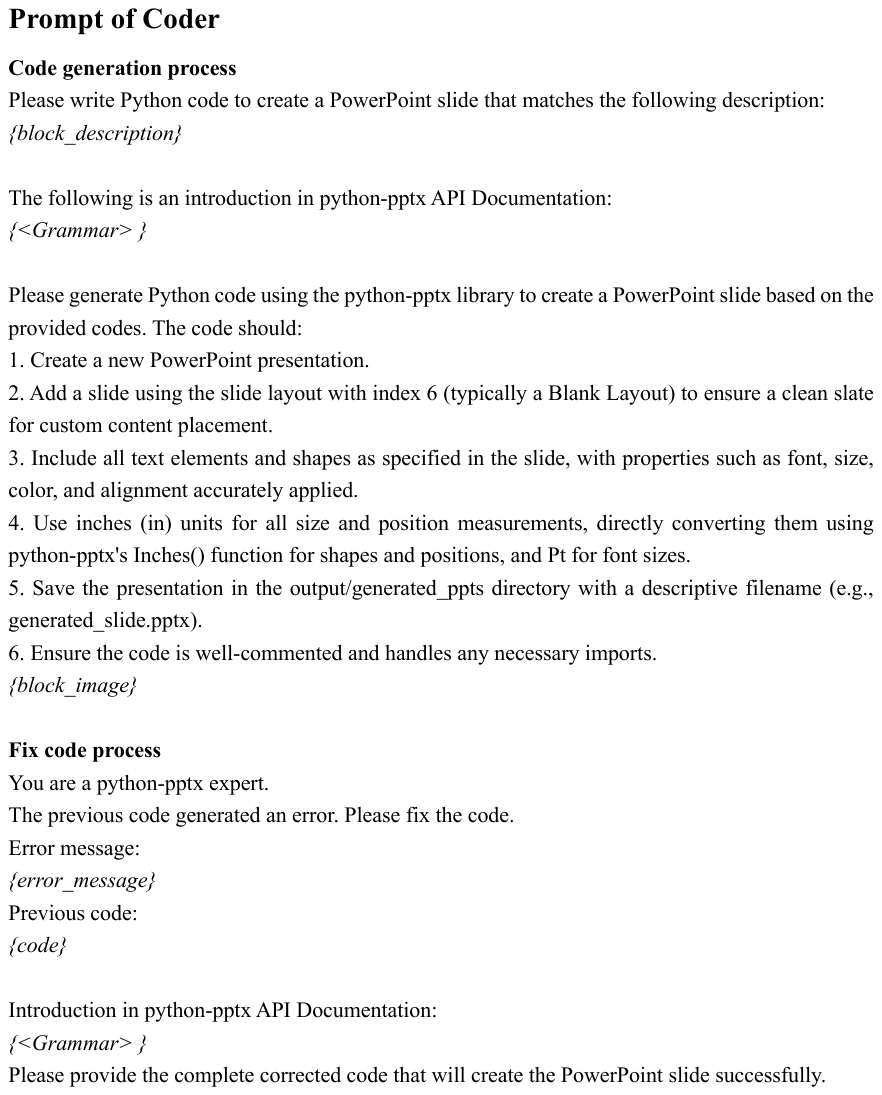 }
  \caption{Prompt of Coder.}
  \label{fig:prompt2}
\end{figure*}

\begin{figure*}[ht]
  \centering
  \includegraphics[width=\textwidth]{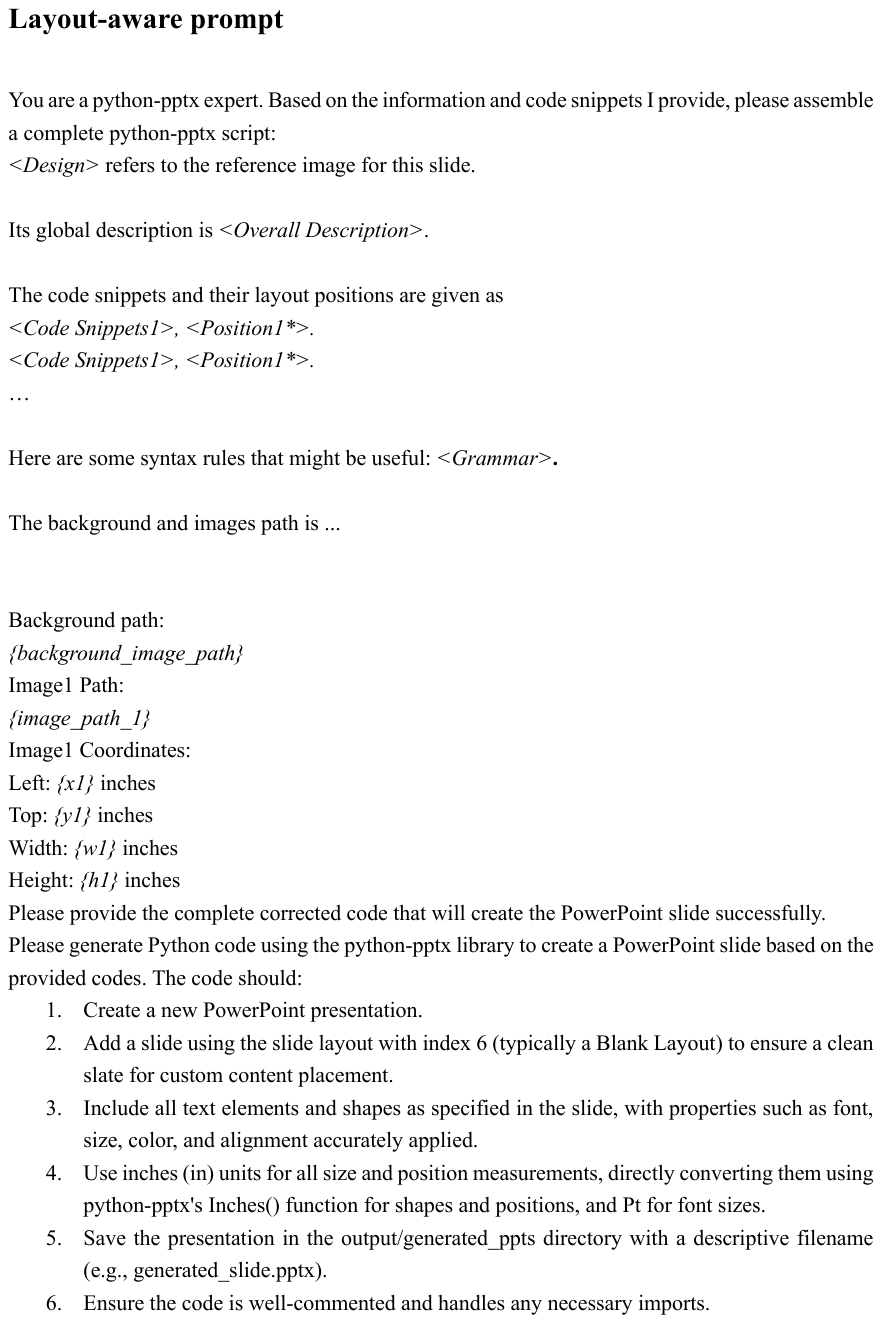}
  \caption{Layout-aware prompt.}
  \label{fig:prompt3}
\end{figure*}

The prompt templates for the Describer and Coder are shown in Figure~\ref{fig:prompt1} and Figure~\ref{fig:prompt2}, respectively. Layout-aware prompt is shown in Figure~\ref{fig:prompt3}.

\section{Details of the Knowledge Base Construction}
\label{sub-app:kb}
\begin{figure*}[ht]
  \centering
  \includegraphics[width=\textwidth]{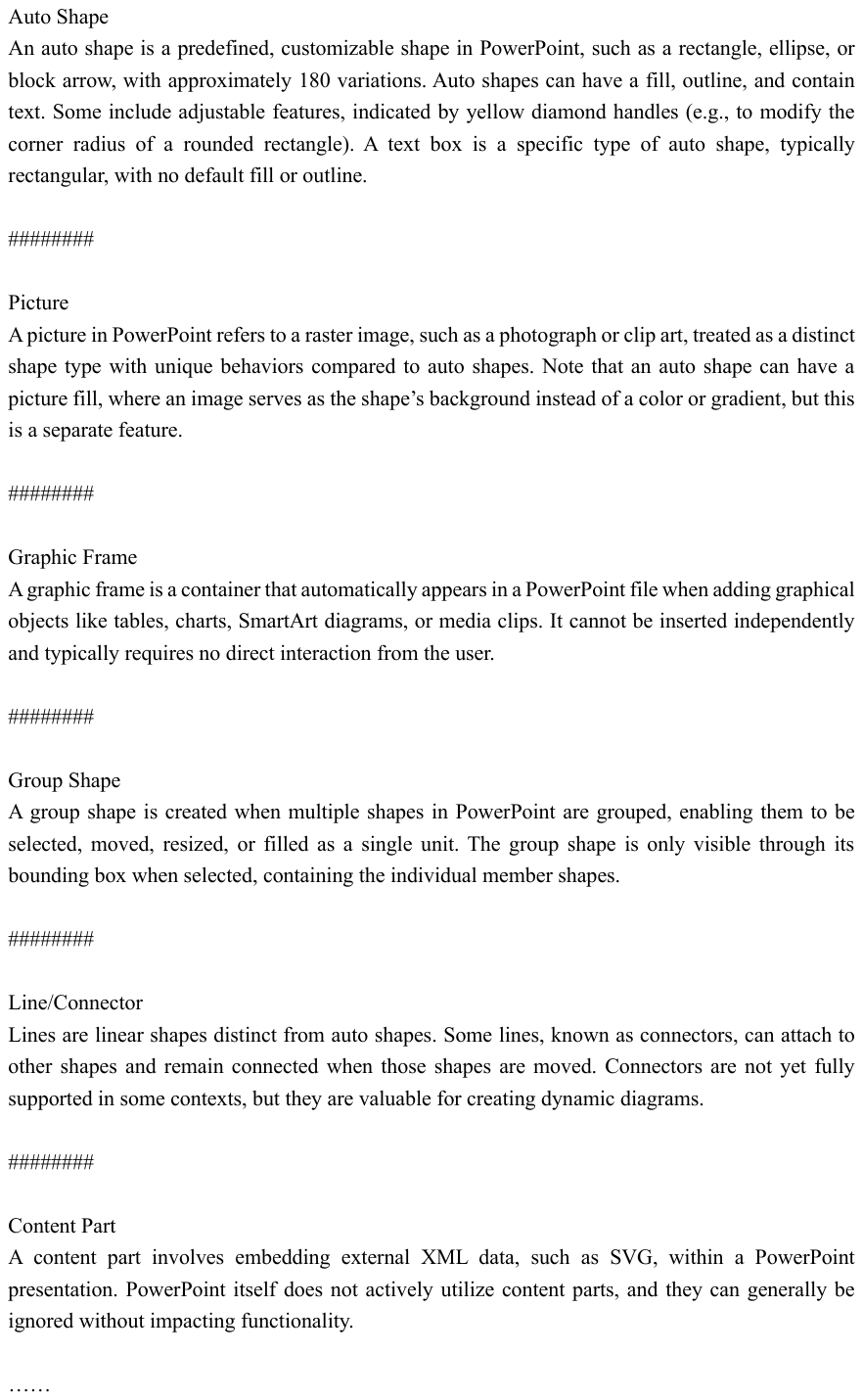}
  \caption{Examples from the Shape Type knowledge base.}
  \label{fig:otkb}
\end{figure*}

\begin{figure*}[ht]
  \centering
  \includegraphics[width=\textwidth]{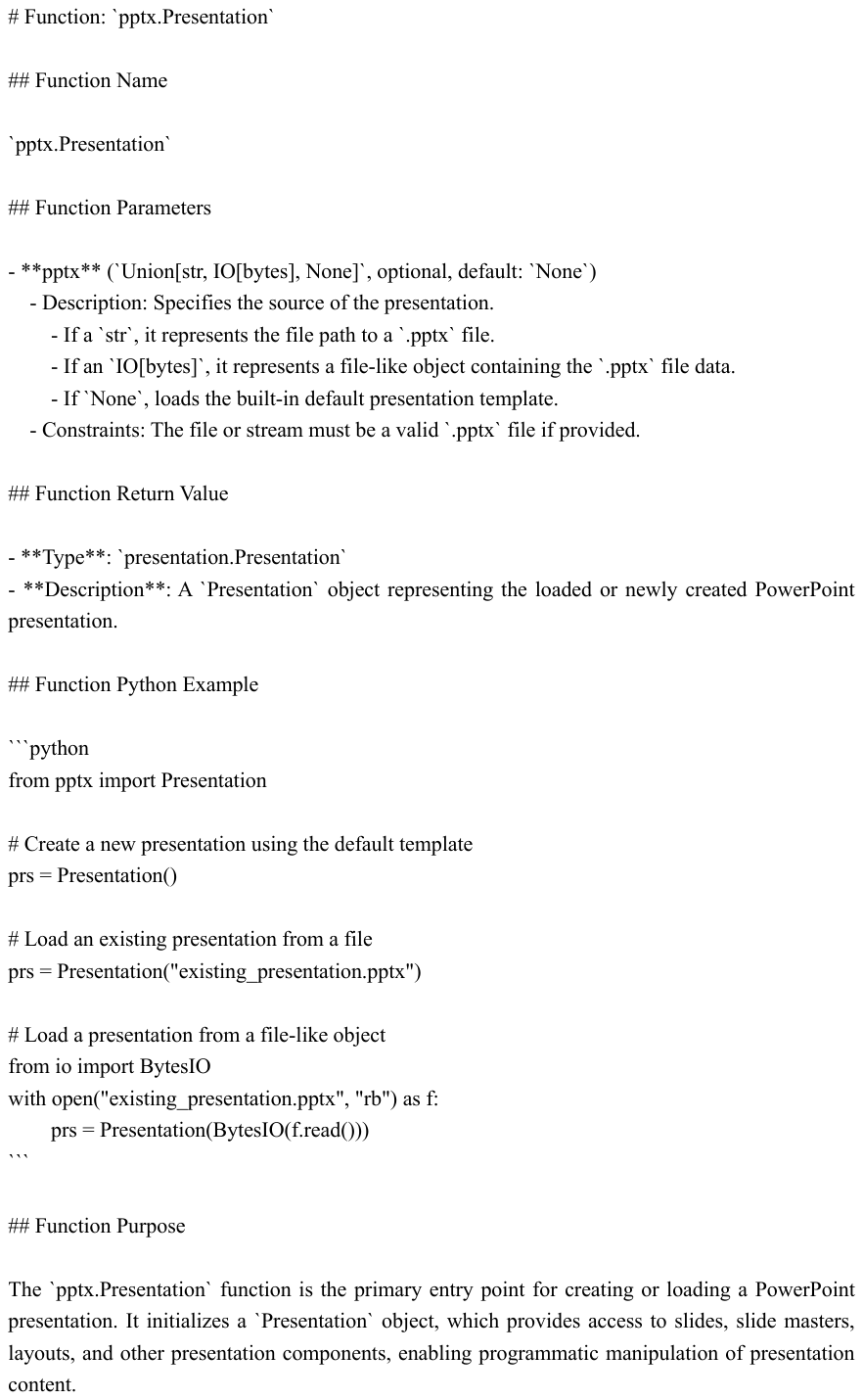}
  \caption{An example from the Operation Function knowledge base.}
  \label{fig:funcKB}
\end{figure*}

Figure~\ref{fig:otkb} presents several examples from the Shape Type Knowledge Base, which consists of object types defined in the python-pptx library along with their corresponding descriptions. Figure~\ref{fig:funcKB} shows an example from the Operation Function Knowledge Base, which includes the function name, parameters, return value, usage example, and a textual explanation of the function.

\begin{table*}[h!]
    \centering
    \caption{Detailed performance analysis under several ablation settings. \colorbox{color11}{Green}, \colorbox{color2}{yellow}, and \colorbox{low}{red} indicate simple, medium, and complex levels in SlideCoder. \textbf{Bolded values} mark the best result per level.}
    \label{tab:detail_ablation}
    \begin{tabular}{c|c|cc|cc|c}
    \toprule
\multirow{2}{*}{\bf{Setting}}&\multirow{2}{*}{\bf{Execution\%}}&\multicolumn{2}{c|}{\bf{Global Visual Metrics}}&\multicolumn{2}{c|}{\bf{Local Structural Metrics}}&\multirow{2}{*}{\bf{Overall}}\\\cline{3-6}
&&\makebox[1.6cm][c]{\textbf{Content}} &
\makebox[1.6cm][c]{\textbf{Position}} &
\makebox[1.6cm][c]{\textbf{Clip}} &
\makebox[1.6cm][c]{\textbf{SSIM}}&\\
\hline
\multirow{3}{*}{SlideCoder}&\cellcolor{color11}\textg{100.0} &\cellcolor{color11}\textg{97.1} &\cellcolor{color11}\textg{89.9} &\cellcolor{color11}80.8 &\cellcolor{color11}\textg{92.9} &\cellcolor{color11}\textg{89.9} \\
&\cellcolor{color2}\texty{100.0} &\cellcolor{color2}92.7 &\cellcolor{color2}\texty{86.5} &\cellcolor{color2}\texty{82.7} &\cellcolor{color2}85.8 &\cellcolor{color2}\texty{85.8}\\ 
&\cellcolor{low}\textr{100.0} &\cellcolor{low}\textr{95.0} &\cellcolor{low}81.3 &\cellcolor{low}\textr{82.2} &\cellcolor{low}82.3 &\cellcolor{low}\textr{82.2} \\\cline{2-7}
\multirow{3}{*}{w/o Layout}&\cellcolor{color11}100.0 &\cellcolor{color11}88.8 &\cellcolor{color11}86.4 &\cellcolor{color11}\textg{81.2} &\cellcolor{color11}79.2 &\cellcolor{color11}81.2 \\
&\cellcolor{color2}93.9 &\cellcolor{color2}90.4 &\cellcolor{color2}75.2 &\cellcolor{color2}80.9 &\cellcolor{color2}78.4 &\cellcolor{color2}73.6 \\
&\cellcolor{low}93.9 &\cellcolor{low}93.6 &\cellcolor{low}72.2 &\cellcolor{low}80.3 &\cellcolor{low}76.4 &\cellcolor{low}71.8 \\\cline{2-7}
\multirow{3}{*}{w/o CGSeg}&\cellcolor{color11}75.8 &\cellcolor{color11}90.4 &\cellcolor{color11}86.5 &\cellcolor{color11}69.4 &\cellcolor{color11}73.1 &\cellcolor{color11}55.4 \\
&\cellcolor{color2}51.5 &\cellcolor{color2}91.7 &\cellcolor{color2}81.4 &\cellcolor{color2}68.5 &\cellcolor{color2}71.4 &\cellcolor{color2}39.6 \\
&\cellcolor{low}69.7 &\cellcolor{low}93.0 &\cellcolor{low}83.2 &\cellcolor{low}68.1 &\cellcolor{low}69.0 &\cellcolor{low}48.4\\\cline{2-7} 
\multirow{3}{*}{w/o H-RAG}&\cellcolor{color11}90.9 &\cellcolor{color11}98.6 &\cellcolor{color11}88.4 &\cellcolor{color11}79.7 &\cellcolor{color11}91.8 &\cellcolor{color11}80.4\\ 
&\cellcolor{color2}81.8 &\cellcolor{color2}91.6 &\cellcolor{color2}84.7 &\cellcolor{color2}81.7 &\cellcolor{color2}\texty{87.8} &\cellcolor{color2}69.3\\
&\cellcolor{low}84.8 &\cellcolor{low}94.0 &\cellcolor{low}\textr{87.9} &\cellcolor{low}81.3 &\cellcolor{low}\textr{83.4} &\cellcolor{low}70.7\\\cline{2-7} 
\multirow{3}{*}{Native Setting}&\cellcolor{color11}75.8 &\cellcolor{color11}90.0 &\cellcolor{color11}87.9 &\cellcolor{color11}71.1 &\cellcolor{color11}71.2 &\cellcolor{color11}53.9\\ 
&\cellcolor{color2}48.5 &\cellcolor{color2}\texty{92.9} &\cellcolor{color2}83.3 &\cellcolor{color2}66.7 &\cellcolor{color2}69.5 &\cellcolor{color2}37.4 \\
&\cellcolor{low}66.7 &\cellcolor{low}92.6 &\cellcolor{low}85.7 &\cellcolor{low}66.5 &\cellcolor{low}70.4 &\cellcolor{low}46.9 

\\
\bottomrule
\end{tabular}
\end{table*}

\begin{table*}[ht]
\centering
\caption{The object types and their styles supported by our reverse engineering tool.}

\label{tab:tool-our}
\begin{tabularx}{\textwidth}{>{\bfseries}l >{\centering\arraybackslash}X}
\toprule
Object Type & \textbf{Styles} \\
\midrule
textbox & Position, Text frame margin, Alignment, Paragraph spacing, Font style, Fill color, Font size, Bold, Italic, Underline \\
rectangle & Position, Line color, Line width, Fill color \\
object\_placeholder & Position, Fill color, Object position \\
freeform & Position, Fill color \\
bullet\_points & Position, Item content, Font size, Font color, Fill color, Bold, Italic, Underline \\
image & Position, Image path \\
background\_color & Color \\
connector & Start position, End position, Arrow color, Arrow width, Arrow style \\
table & Position, Cell height, Cell fill color, Text inside cell \\
triangle & Position, Type, Line color, Line width, Fill color \\
\bottomrule
\end{tabularx}
\label{tab:graphical-elements}
\end{table*}

\begin{table*}[ht]
\centering
\caption{The object types and their styles supported by AutoPresent's reverse engineering tool.}

\label{tab:tool-ap}
\begin{tabularx}{\textwidth}{>{\bfseries}l >{\centering\arraybackslash}X}
\toprule
Object Type & \textbf{Styles} \\
\midrule
title & Font size, Font color, Fill color \\
textbox & Position, Font size, Bold, Font color, Fill color \\
bullet\_points & Position, Item content, Font size, Font color, Fill color \\
image & Position, Image path \\
background color & Color \\
\bottomrule
\end{tabularx}
\label{tab:slide-structure}
\end{table*}

\begin{table}[t]
\centering
\caption{LoRA fine-tuning configuration used in our experiments.}
\label{tab:lora-config}
\begin{tabular}{ll}
\toprule
\textbf{Parameter} & \textbf{Value} \\
\midrule
Rank & 8 \\
Max Sequence Length & 4096 \\
Batch Size & 4 \\
Gradient Accumulation Steps & 8 \\
Learning rate & 1e-4 \\
Epochs & 10 \\
Warmup Ratio & 0.1 \\
Mixed Precision & bf16 \\
\bottomrule
\end{tabular}
\end{table}

\end{document}